\documentclass{edm_template}

\usepackage{algpseudocode}
\usepackage{algorithm}
\usepackage[algo2e]{algorithm2e}
\usepackage{amsmath}
\usepackage{booktabs}
\usepackage[scaled=0.75]{beramono}
\usepackage{xspace}
\usepackage{multirow}
\usepackage{arydshln}
\usepackage[hyphens]{url}
\usepackage{bbm}
\usepackage{adjustbox}
\usepackage{graphicx}
\usepackage{comment}
\usepackage{enumitem}

\newcommand{\dataset}[1]{\texttt{#1}\xspace}
\newcommand{\bank}{\dataset{bank}}
\newcommand{\pharma}{\dataset{pharma}}

\newcommand{\method}[1]{\textsc{#1}\xspace}
\newcommand{\syntactic}{\method{syntactic-chunker}}
\newcommand{\keyphrase}{\method{keyphrase}}
\newcommand{\learningobjective}{\method{keyphrase Reranking}}
\newcommand{\bloom}{\method{bloom-verb}}
\newcommand{\semantic}{\method{semantic-chunker}}
\newcommand{\hybrid}{\method{hybrid-chunker}}
\newcommand{\textrank}{\method{Modified TextRank \cite{IBMJournalPaper}}}
\newcommand{\alchemy}{\method{Watson NLU}}
\newcommand{\wordtovec}{\method{word2vec}}

\newcommand{\secref}[2][]{Section#1~\ref{sec:#2}}

\newcommand{\tabref}[2][]{Table#1~\ref{tab:#2}}
\newcommand{\figref}[2][]{Figure#1~\ref{fig:#2}}

\newcommand{\algoref}[2][]{Algorithm#1~\ref{alg:#2}}

\makeatletter
\newcommand{\removelatexerror}{\let\@latex@error\@gobble}
\makeatother

\begin{document}
\title{Document Chunking and Learning Objective Generation for Instruction Design} 

\author{Khoi-Nguyen Tran\\ \affaddr{IBM Research}\\ \affaddr{ Australia} \\ \email{\small khndtran@au1.ibm.com} \\ 
\and
Jey Han Lau\\ \affaddr{IBM Research}\\ \affaddr{Australia} \\ \email{\small jeyhan.lau@au1.ibm.com} \\ 
\and
Danish Contractor\\ \affaddr{IBM Research}\\ \affaddr{India}\\ \email{\small dcontrac@in.ibm.com} \\  
\and
Utkarsh Gupta\titlenote{Utkarsh carried out this work during his employment with IBM Research.}\\ \affaddr{IBM Research}\\ \affaddr{India}\\ \email{\small utgupta3@in.ibm.com} \\ 
\and
Bikram Sengupta\\ \affaddr{IBM Research}\\ \affaddr{India}\\ \email{\small bsengupt@in.ibm.com} \\ 
\and
Christopher J. Butler \\ \affaddr{IBM Research}\\ \affaddr{Australia}\\ \email{\small chris.butler@au1.ibm.com} \\ 
\and
Mukesh Mohania\\ \affaddr{IBM Research}\\ \affaddr{Australia}\\ \email{\small mukeshm@au1.ibm.com} \\ 
}

\date{}
\maketitle

\begin{abstract}
Instructional Systems Design is the practice of creating of instructional experiences that make the acquisition of knowledge and skill more efficient, effective, and appealing~\cite{ID}.
Specifically in designing courses, an hour of training material can require between 30 to 500 hours of effort in sourcing and organizing reference data for use in just the preparation of course material.
In this paper, we present the first system of its kind that helps reduce the effort associated with sourcing reference material and course creation. We present algorithms for document chunking and automatic generation of learning objectives from content, creating descriptive content metadata to improve content-discoverability. Unlike existing methods, the learning objectives generated by our system incorporate pedagogically motivated Bloom's verbs. We demonstrate the usefulness of our methods using real world data from the banking industry and through a live deployment at a large pharmaceutical company.


\end{abstract}

%
%



\section{Introduction}


Recent estimates suggest that on average, an organization spends nearly $\$1200$ per year, per employee for training.\footnote{\url{https://www.td.org/Publications/Magazines/TD/TD-Archive/2014/11/2014-State-of-the-Industry-Report-Spending-on-Employee-Training-Remains-a-Priority}.} Apart from the costs incurred in delivering training, significant costs are associated with instruction design activities such as sourcing and preparation of course materials. Currently, most of these activities are very human-intensive in nature, and they rely on the experience and expertise levels of instruction designers and intense reviews by subject-matter experts (SMEs) to achieve acceptable quality levels. 

\subsection{Course Creation: Workflow and Challenges}

\begin{figure}[!t]
\centering
\includegraphics[scale=0.55]{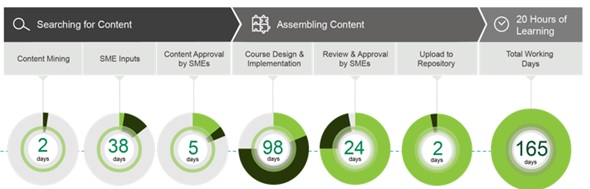}
\caption{Typical course creation workflow}\label{fig:ID-flow}
\end{figure}


\figref{ID-flow} shows the typical steps involved in creating a new course. In the first step, instructional designers search for existing learning content that can be used for reference while developing the course. The learning objectives of the new (to be designed) course informs this search process. Reference materials may include existing courses and resources as well as other informal learning materials, such as those available in the form of media articles, blogs etc. 

In the next step, the new course is designed and implemented by: extracting the relevant parts of the selected reference content, transforming them appropriately, and combining with newly developed materials to meet the overall training objectives. The new course content is finalized with SME review and approval. Finally, the course is uploaded to a repository for access by end users such as instructors and employees.
 
The average time taken to produce an hour of material this way can vary between 50 to 300 hours depending on the nature of the course being created.\footnote{\url{https://www.td.org/Publications/Newsletters/Learning-Circuits/Learning-Circuits-Archives/2009/08/Time-to-Develop-One-Hour-of-Training}.} The efficiency with which a new course can be assembled rests on two critical factors: (a) the ability to quickly locate an existing reference material, which is relevant to a learning objective that is part of the planned new course; and (b) the ability to identify (and eventually extract) appropriate parts of this material for use within the new course. 

\subsection{Contributions}
In this paper, we present the first system of its kind that helps reduce the effort associated with sourcing reference material and course creation. We present algorithms for document chunking and automatically generating learning objectives from content as well as  creating descriptive content meta-data that improves content-discoverability. Our novel methods for document chunking incorporate syntactic and stylistic features from text as well as a semantic vector-based representation of document text to identify meaningful chunks. Each chunk is physically persisted and a learning objective consisting of Bloom's verb~\cite{Bloom1984} along with a descriptive keyphrase is generated and associated with each chunk. To the best of our knowledge, we are the first to generate learning objectives incorporating Bloom's verbs and our system is the first of its kind that directly addresses the challenges in instruction design. 


We describe experiments using real-world data from two industries: banking and pharmaceutical. Our results on data from the banking industry shows that our document chunking methods are useful for instruction designers. We report an average user rating of $2$ out of $3$ in a blind study to assess the quality of chunks and an $F1$ score of $0.62$ computed against expert generated gold standard chunks. Furthermore, in the challenging problem of generating learning objectives, the output from our system has an $F1$ score of $0.70$ for predicting Bloom's verbs with an average user rating of $2.2$ (out of $3$) for the associated keyphrase. We also present details of a live deployment of our solution at a large pharmaceutical company.

\section{Related work} \label{sec:related}

To the best of our knowledge, our system is the first (commercial or prototype) that can automatically chunk/segment\footnote{We use the word ``chunk'' and ``segment'' interchangeably, though a document chunk further refers to a physical embodiment of a document segment}
learning material and label them with system-generated course objectives. We highlight some related work directly relevant to the subcomponents of document chunking and learning objective generation.

\noindent \textbf{Document chunking:} Broadly, most methods for chunking/segmentation of text rely on detecting changes in vocabulary usage patterns \cite{hearst1993texttiling,kazantseva2012topical,kazantseva2014measuring}, identifying topical shifts \cite{du2013topic,du2015topic,seg1}, or employing graph based techniques to identify boundaries \cite{graph-top,tagarelli2013segment}. The TextTiling \cite{hearst1993texttiling} document segmentation algorithm uses shifts in vocabulary patterns to mark segment boundaries. Works such as Riedl and Biemann~\cite{riedl2012topictiling} adapt the TextTiling algorithm to work on topics generated by Latent Dirichlet Allocation. Glavis et al.\cite{graph-top} use a graph based representation of documents based on semantic relatedness of sentences to identify document segments.
More recent work \cite{alemi2015text,EDMpaper} uses semantic distance computed based on vector embeddings to identify chunk/segment boundaries. Our work on document chunking is based on this direction of research. We use file format specific APIs to physically persist document chunks, retaining any stylistic and presentation elements from the original document. 

\noindent \textbf{Learning Objective generation:} Most learning management solutions either rely on user provided learning objectives or automated methods to label documents with \textit{existing} learning objectives specified in curricula \cite{IBMJournalPaper}. Methods such as Bhartiya et al.~\cite{EDMpaper} and Contractor et al.~\cite{SDM2015} use a curriculum hierarchy to label learning material with learning objectives. Milli and Hearst~\cite{BEA2016} simplify the problem of generating course objectives by directly using document keyphrases as learning objectives. Similarly, Lang et al.~\cite{Lang2018} and Rouly et al.~\cite{Rouly2015} simplify generating objectives using topic modeling to identify candidate learning objectives, where Lang et al.~\cite{Lang2018} also suggest a system to match topics with Bloom's verbs. In contrast, we associate keyphrases with Bloom's verbs~\cite{Bloom1984} and rerank them to select the best candidates for use as learning objectives. To the best of our knowledge, we are the first to \textit{generate} pedagogically motivated learning objectives incorporating Bloom's verbs.


\section{Document Chunking}\label{sec:chunking}

Course materials can often be very large and monolithic, covering a great number of topics and learning objectives, which makes consumption difficult.
To make these course materials more discoverable, we automatically segment courses into smaller chunks that can persist independently in the course repository.
We present three chunking approaches in the following sections.











\subsection{Structure guided (\syntactic)}

Section headings are often the most natural chunk boundaries as they reflect the organization of content by the document creator. Formats such as Microsoft Word have an underlying XML structure that allows us to create these natural chunks easily. However, for PDF documents, there is no encoded document structure information, but we can recover the section titles by analyzing the font sizes of text. To build the \syntactic, we use a combination of Apache PDFBox\footnote{https://pdfbox.apache.org/} for PDF documents, Aspose APIs\footnote{https://docs.aspose.com/dashboard.action} for Microsoft Office documents and Apache Tika\footnote{https://tika.apache.org/} for all other document formats.

Algorithm~\ref{alg:syntactic} details the syntactic chunking algorithm where we do not have markers for the section headings.
The algorithm aims to find the font size of the largest heading in the document for chunking. The \syntactic first groups the lines in the document by their font size (sequentially). 
For each of these font groups, the algorithm gathers statistics on the chunks that would be created for each group's font size.
The largest font size (i.e. the top most section titles) is then chosen from the groups that satisfies the heuristics given in the chunking hyperparameters. An example heuristic is whether the number of chunks created by this font size is between $3$ and $20$, which is the number of sections or subsections we expect a document or a chapter to contain on average. The significant heuristics/hyperparameters for this algorithm are given in \tabref{hyperparams:syntactic}.

Finally, the line indices marking the start of the section headings are recovered through the font groups created earlier. These starting line indices are then further processed in the main algorithm for creating the physical chunks or storing the metadata. 

\RestyleAlgo{algoruled}
\LinesNumbered

\begin{figure}[!t]
\removelatexerror
\begin{algorithm2e}[H]
\DontPrintSemicolon

\SetKwData{PDF}{pdf}
\SetKwData{lineText}{lineText}
\SetKwData{lineFS}{lineFS}
\SetKwData{fg}{fg}
\SetKwData{fgs}{fgs}
\SetKwData{cStats}{cStats}
\SetKwData{cs}{cs}
\SetKwData{chunkingFontSize}{chunkingFontSize}
\SetKwData{chunkStartIndices}{chunkStartIndices}

\SetKwFunction{LoadDocument}{LoadDocument}
\SetKwFunction{LoadParameters}{LoadParameters}
\SetKwFunction{ExtractOnEachLine}{ExtractOnEachLine}
\SetKwFunction{GetStats}{GetStats}
\SetKwFunction{Heuristics}{Heuristics}
\SetKwFunction{LargestFontSize}{LargestFontSize}

\SetKwInOut{Input}{Input}
\SetKwInOut{Output}{Output}

\Input{A path to the document}
\Output{A list of indices to lines/pages in the document marking the start of a chunk}

\LoadParameters{``$syntactic$''}\;
\PDF $\leftarrow$ \LoadDocument{}\;
\lineText $\leftarrow$ \ExtractOnEachLine{``$text$'', \PDF}\;
\lineFS $\leftarrow$ \ExtractOnEachLine{``$fontsize$'', \PDF}\;

\tcp{Font groups are contiguous groups of lines.}
\fgs $\leftarrow$ $[(i, k-1)\ |\ \lineFS[i] = \lineFS[j], i \leq j < k ]$\;

\tcp{Create chunk statistics for each font group}
\For{$i, j \in \fgs.length,\ i = j$}{
	\While{$\lineFS[\fgs[i]] \geq \lineFS[\fgs[j]]$}{
    	$\cStats[\lineFS[\fgs[i]]] \mathrel{+}=$ \GetStats{$\fgs\textup{[}j\textup{]}$}\;
        $j \leftarrow j+1\;$
	}
}

\tcp{Select candidates from heuristics}
\cs $\leftarrow$ $[\fg\ |\ $\Heuristics{$\fg, \cStats\textup{[}\fg\textup{]}$}$, \forall \fg \in \fgs]$\;

\chunkingFontSize $\leftarrow$ \LargestFontSize{\cs}\;

\tcp{Return the chunk start boundaries}
\chunkStartIndices $\leftarrow$ $[\fg.startIndex\ |\ \lineFS[\fg] = \chunkingFontSize, \forall \fg \in \fgs]$\;

\caption{Syntactic chunking algorithm}
\label{alg:syntactic}
\end{algorithm2e}
\end{figure}

\begin{table}[!t]
	\footnotesize
 	\begin{tabular}{ p{2.5cm} p{1cm} p{3.8cm}}
     \toprule
 	\textbf{Hyperparameter}  & \textbf{Value} & \textbf{Description}\\
 	\midrule
 	\texttt{font\_group\_lines} & [1,3] & Minimum and maximum number of consecutive lines (of the same font size) to collapse. \\
 	\texttt{n\_chunks} & [3, 20] & Minimum and maximum number of resulting chunks for each font size. \\
 	\texttt{min\_section\_ title\_length} & 2 & Minimum number of characters for a chunk's starting line. \\
 	\bottomrule
 	\end{tabular}
 	\caption{Syntactic-chunker hyperparameters.}
 	\label{tab:hyperparams:syntactic}
\end{table}

\begin{table}[!t]
	\footnotesize
	\begin{tabular}{ p{2.5cm} p{1cm} p{3.8cm}}
	\toprule
	\textbf{Hyperparameter} & \textbf{Value} & \textbf{Description}\\
	\midrule
	\texttt{min\_par\_to\_stop} & 80 & Threshold for the minimum number of lines to stop chunking. \\
	\texttt{trim\_par} & 4 & Proportion of starting and ending lines to ignore when searching for a chunk boundary. \\
	\texttt{word2vec\_model} & enwiki & Pre-trained \wordtovec model. \\
	\texttt{max\_vocab} & 1000 & Number of most frequent word types to include from pre-trained \wordtovec model. \\
	\bottomrule
	\end{tabular}
	\caption{Semantic-chunker hyperparameters.}
	\label{tab:hyperparams:semantic}
\end{table}

\RestyleAlgo{algoruled}
\LinesNumbered

\begin{figure}[!t]
\removelatexerror
\begin{algorithm2e}[H]
\footnotesize{

\DontPrintSemicolon

\SetKwData{PDF}{pdf}
\SetKwData{lineText}{lineText}
\SetKwData{lineFS}{lineFS}
\SetKwData{fg}{fg}
\SetKwData{fgs}{fgs}
\SetKwData{fgsV}{fgsV}
\SetKwData{cStats}{cStats}
\SetKwData{cs}{cs}
\SetKwData{chunkingFontSize}{chunkingFontSize}
\SetKwData{chunkStartIndices}{chunkStartIndices}
\SetKwData{lineVectors}{lineVectors}
\SetKwData{startIndex}{startIndex}
\SetKwData{numParagraphsInChunk}{numParagraphsInChunk}
\SetKwData{minNumberOfLines}{minNumberOfLines}

\SetKwFunction{LoadDocument}{LoadDocument}
\SetKwFunction{LoadParameters}{LoadParameters}
\SetKwFunction{ExtractOnEachLine}{ExtractOnEachLine}
\SetKwFunction{GetStats}{GetStats}
\SetKwFunction{Heuristics}{Heuristics}
\SetKwFunction{LargestFontSize}{LargestFontSize}
\SetKwFunction{Vectorize}{Vectorize}
\SetKwFunction{VectorSum}{VectorSum}
\SetKwFunction{VectorSubtract}{VectorSubtract}

\SetKwInOut{Input}{Input}
\SetKwInOut{Output}{Output}

\SetKwProg{Fn}{Function}{:}{}
\SetKwFunction{FindSegments}{FindSegments}
\SetKwFunction{Size}{Size}
\SetKwFunction{Cosine}{Cosine}

\SetKwData{chunkIndices}{chunkIndices}
\SetKwData{bestIndex}{bestIndex}
\SetKwData{bestScore}{bestScore}
\SetKwData{x}{x}
\SetKwData{y}{y}
\SetKwData{n}{n}
\SetKwData{sumTop}{sumTop}
\SetKwData{sumBot}{sumBot}
\SetKwData{cos}{cos}
\SetKwData{topVectors}{topVectors}
\SetKwData{botVectors}{botVectors}

\Input{A path to the document}
\Output{A list of indices to lines/pages in the document marking the start of a chunk}

\LoadParameters{``$semantic$''/``$hybrid$''}\;
\PDF $\leftarrow$ \LoadDocument{}\;
\lineText $\leftarrow$ \ExtractOnEachLine{``$text$'', \PDF}\;
\tcp{Vectorize words using pre-trained word vectors}
\lineVectors $\leftarrow$ $\Vectorize(\lineText)$

\BlankLine
\tcc{Modifications for the hybrid algorithm}
\lineFS $\leftarrow$ \ExtractOnEachLine{``$fontsize$'', \PDF}\;
\tcp{Create font groups.}
\fgs $\leftarrow$ $[(i, k-1)\ |\ \lineFS[i] \equiv \lineFS[j], i \leq j < k ]$\;
\tcp{Vectorize the font groups}
\fgsV $\leftarrow$ $[\VectorSum(\Vectorize(\forall\lineText \in fg))\ |\ \forall\fg \in \fgs]$\;
\tcp{Similar logic to the semantic algorithm}
\lineVectors $\leftarrow$ \fgsV\;

\BlankLine

\tcp{Return the chunk start boundaries (function below)}
\chunkStartIndices $\leftarrow$ \FindSegments{\lineVectors, \startIndex}\;

\BlankLine
\tcc{Divide and conquer strategy}

\Fn{\FindSegments{\lineVectors, \startIndex}}{

\n $\leftarrow$ $\Size(\lineVectors)$\;

\tcp{Create the search area with the \numParagraphsInChunk hyperparameter}
\x $\leftarrow$ $\n/\numParagraphsInChunk$\;
\y $\leftarrow$ $\n/(1-(1/\numParagraphsInChunk))$\;
\bestIndex $\leftarrow$ $(x+y)/2$\;
\bestScore $\leftarrow$ $1.0$\;
\sumTop $\leftarrow$ $\VectorSum(\lineVectors[1,x])$\;
\sumBot $\leftarrow$ $\VectorSum(\lineVectors[x+1, \n])$\;

\BlankLine
\For{$x \leq i < y$}{
	\sumTop $\leftarrow$ $\VectorSum(\sumTop, \lineVectors[i])$\;
    \sumBot $\leftarrow$ $\VectorSubtract(\sumBot, \n)$\;
    \cos $\leftarrow$ \Cosine{sumTop, sumBot}\;
    \If{\cos $<$ \bestScore}{
    	\bestIndex $\leftarrow$ $i$\;
        \bestScore $\leftarrow$ \cos\;
    }
    \chunkIndices.append([\bestIndex $+$ \startIndex])\;
    \topVectors $\leftarrow$ $\lineVectors[1, \bestIndex]$\;
    \botVectors $\leftarrow$ $\lineVectors[\bestIndex+1, \n]$\;
    \BlankLine
    \tcp{Hyperparameter \minNumberOfLines as the stopping condition}
    \If{$\Size(\topVectors) > \minNumberOfLines$} {
    	\chunkIndices.appendAll(\FindSegments{\topVectors, \startIndex})\;
    }
    \If{$\Size(\botVectors) > \minNumberOfLines$}{
    	\chunkIndices.appendAll(\FindSegments{\botVectors, \bestIndex $+$ \startIndex})\;
    }
}
\Return{\chunkIndices}
}
}
\caption{Semantic/hybrid chunking algorithm}
\label{alg:semantic}
\end{algorithm2e}
\end{figure}

\subsection{Topically guided (\semantic)} %

Some document styles have ambiguous semantic separation of content, such as presentation slides, informal articles, and blogs. These document styles often have repeated font sizes and text that do not provide distinguishing characteristics for syntactic chunking. For example, presentation slides often have repeated font sizes for slide titles, causing the \syntactic to create a separate chunk for each slide. For these documents, their text content is more useful for inferring chunk boundaries than syntactic markers.



To chunk these documents, we use a divide-and-conquer approach based on topical or content shifts.  We represent the content using mean bag-of-word embeddings, which are pre-trained \wordtovec embeddings \cite{Mikolov+:2013a,Mikolov+:2013b}.\footnote{Word embeddings are trained on English Wikipedia.} We tokenise words using whitespace, and discard common symbols such as commas and periods. When computing the mean embedding, stopwords are excluded.\footnote{We use mallet's stopword list: \url{https://github.com/mimno/Mallet/blob/master/stoplists/en.txt}}
The divide-and-conquer method first identifies a boundary that separates a document into two partitions that have the maximum cosine distance using the vector embeddings (providing topical diversity), and then recursively creates subpartitions until a minimum text length is reached. The search strategy is simpler compared to dynamic programming and iterative improvement techniques typically used in the literature \cite{alemi2015text} but we found this divide-and-conquer strategy performs encouragingly.   

The pseudocode and hyperparameters for the \semantic algorithm with modifications to create the \hybrid are in \algoref{semantic} and \tabref{hyperparams:semantic}, respectively. Both algorithms share similar hyperparameters and similar divide-and-conquer logic but on different data structures.


\subsection{Hybrid method (\hybrid)}

The \semantic relies purely on content information for chunking, ignoring potentially usable structural information. From preliminary experiments, we observed that the \semantic occasionally partitions documents at arbitrary positions in the text. For example, a few lines after the start of a new section where the topical shift should be stronger. To resolve this, we developed a hybrid method that uses both structural and content information. Similar to the \syntactic, we record font sizes for each line, and gather lines that share a similar font size into a data structure. With these data structures, we apply the same divide-and-conquer approach used in the \semantic to recursively partition the document into multiple chunks. This forces the chunker to create partitions at natural text boundaries, when this information is available.

\section{Learning Objective Generation}\label{sec:lo}

Traditionally, learning objectives associated with courses are generated manually and are presented in a sentence-like structure. An example from a K-12 Science curriculum in the US:
\textit{Conduct an investigation to determine whether the mixing of two or more substances results in new substances.}\footnote{Sources: \url{https://www.cs.ox.ac.uk/teaching/courses/2015-2016/ml/}, \url{https://www.nextgenscience.org/topic-arrangement/5structure-and-properties-matter}.}

Automatically generating these objectives can be posed as summarization problem where the task is to identify the ``\textit{learning skill}'' imparted by the document. However, inferring a skill requires an in-depth understanding of the concepts presented, how they relate with each other, and in courses--such as those that teach soft-skills or behavioural skills--the relationships may be more abstract. Thus, in order to generate tractable yet usable learning objectives, we generate short sentences that are prefixed by a verb from the Bloom's taxonomy followed by a keyphrase. Recent work such as Milli and Hearst~\cite{BEA2016} contends with simply using keyphrases as learning objectives.



\subsection{Candidate Keyphrase Selection}
Existing methods for keyphrase extraction use a variety of different approaches. Some methods rely on supervision to extract keyphrases \cite{Jiang+:2009,Song+:2003,Witten+:1999}, while unsupervised methods often rely on graph-based ranking \cite{TextRank} or topic-based clustering \cite{Grineva+:2009,Liu+:2009}.
For our work, we rely on an accessible and effective keyphrase extraction method: IBM Watson Natural Language Understanding (NLU)\footnote{\url{https://natural-language-understanding-demo.mybluemix.net/}} to extract keyphrases. NLU is one of many commercially available general purpose keyphrase extraction methods that performs effectively in general keyphrase extraction tasks~\cite{alchemy1,alchemy2}. We also evaluated other methods such as a variant of TextRank~\cite{TextRank}, which has been used in extracting keyphrases from education material~\cite{IBMJournalPaper}.
We chose NLU for the rest of this paper after a blind user study on $243$ document chunks indicated a strong preference for these keyphrases as compared to the method employed by Contractor et al.~\cite{IBMJournalPaper}.
\tabref{kp-method} shows the proportion of useful keyphrases\footnote{``Usefulness'' is defined in terms of possible inclusion of a keyphrase in a learning objective, and not in terms of the ``quality'' of a keyphrase in a general keyphrase extraction task.} for two keyphrase extraction methods. Further details and results are given in \secref{exp:lo}.




\begin{table}[!t]
\footnotesize
\begin{center}
\begin{tabular}{cc}
\toprule
\textbf{Method} & \textbf{\% {Useful} Keyphrases}\\
\midrule
\alchemy & 66 \\
\textrank & 51 \\
\bottomrule
\end{tabular}
\end{center}
\caption{Percentage proportion of keyphrases identified by instructional designers as being ``{\em useful}'' for possible inclusion in learning objectives}
\label{tab:kp-method}
\end{table}

As seen from \tabref{kp-method}, not all keyphrases extracted are useful for inclusion in learning objectives. Thus, to select candidate keyphrases for learning objectives from a general keyphrase list, we rank and select them using a combination of factors:

\begin{enumerate}[itemsep=0mm, leftmargin=*]

\item {\bf Keyphrase score ($\alpha$):} A score between 0-1 returned by the NLU indicating the importance of a keyphrase (1 = most important). 

\item {\bf N-gram TF-IDF score ($\beta$):} We compute an N-gram level TF-IDF score for each keyphrase using a large domain specific background corpus for IDF score computation.

\item {\bf Inverse chunk frequency ($\gamma$):} We compute a chunk-level  modified IDF score for each keyphrase where the IDF score is computed at the keyphrase level using sibling chunks of a given chunk. 

\item {\bf Google N-gram score ($\phi$):} The Google Books N-gram service\footnote{\url{https://books.google.com/ngrams}.} returns the log-likelihood of a given N-gram from a language model trained on the Google Books corpus. 
We use the (normalized) rank for a keyphrase within a chunk as the N-gram score.

\item {\bf Word token level overlap with document section titles  ($\theta$):} Tokens in a section title are likely to contain mentions of important concepts  and this acts as a useful signal for selecting keyphrases for learning objectives.

\end{enumerate}

Let weight $w_i$ be associated with each scoring factor $f_i$, where there are $N$ factors. The weights of each factor is normalized to sum to 1.0 (i.e.\ $\sum_{i=0}^{N}{w_i}=1.0$). Let $K_{s}^{(j)}$ denote the set of top-$k$ keyphrases selected by the system for the $j$-th chunk (based on decreasing order of the score $\sum_{i=0}^{N}{w_i}{f_i}$).  Let the average user rating (see Section~\ref{sec:exp:lo}) associated with the keyphrase set $K_{s}^{(j)}$ be denoted by $s_j$. 
Our goal is to select values of $w_i$ that maximises $s_j$ for all training examples:
\begin{equation}
max \frac{\sum_{j=1}^{M} s_j}{M}
\label{eq:maxKPAverage}
\end{equation}
where $M$ is the number of training examples. The values for $k$ and parameters $w_i$ are estimated using grid search.

The tuned hyperparameters for the keyphrase selection are given in \tabref{hyperparams:keyphrase}. We found that $\mathbf{\gamma}$ is not useful in these data sets, but maybe useful in other document collections where learning objectives are derived from a few chunks.

\begin{table}[t]
	\centering
    \begin{tabular}{rccccc}
    \toprule
     & $\mathbf{\alpha}$ & $\mathbf{\beta}$ & $\mathbf{\gamma}$ & $\mathbf{\phi}$ & $\mathbf{\theta}$ \\
    \midrule
    \bank & 0 & 0.5 & 0 & 0.5 & 0 \\
    \pharma & 0.26 & 0.32 & 0 & 0.32 & 0.1 \\
    \bottomrule
    \end{tabular}
	\caption{Hyperparameter values for \bank and \pharma data for keyphrase re-ranking: $\alpha$: orignal keyphrase score, $\beta$: N-gram TF-IDF score, $\gamma$: Inverse Chunk Frequency, $\phi$: Google N-gram score, $\theta$: Overlap with words in section titles.}
	\label{tab:hyperparams:keyphrase}
\end{table}

%






\subsection{Bloom's Verbs Association}

Bloom~\cite{Bloom1984} proposes a taxonomy for promoting learning instead of rote memorization. Bloom's taxonomy aims to capture the whole pedagogy of learning, teaching, and processing information in a list of ``{action}'' verbs. 
These verbs (referred to as Bloom's verbs) characterize the activity involved in learning concepts. 

\begin{figure}[!t]
\centering
\includegraphics[width=0.5\textwidth]{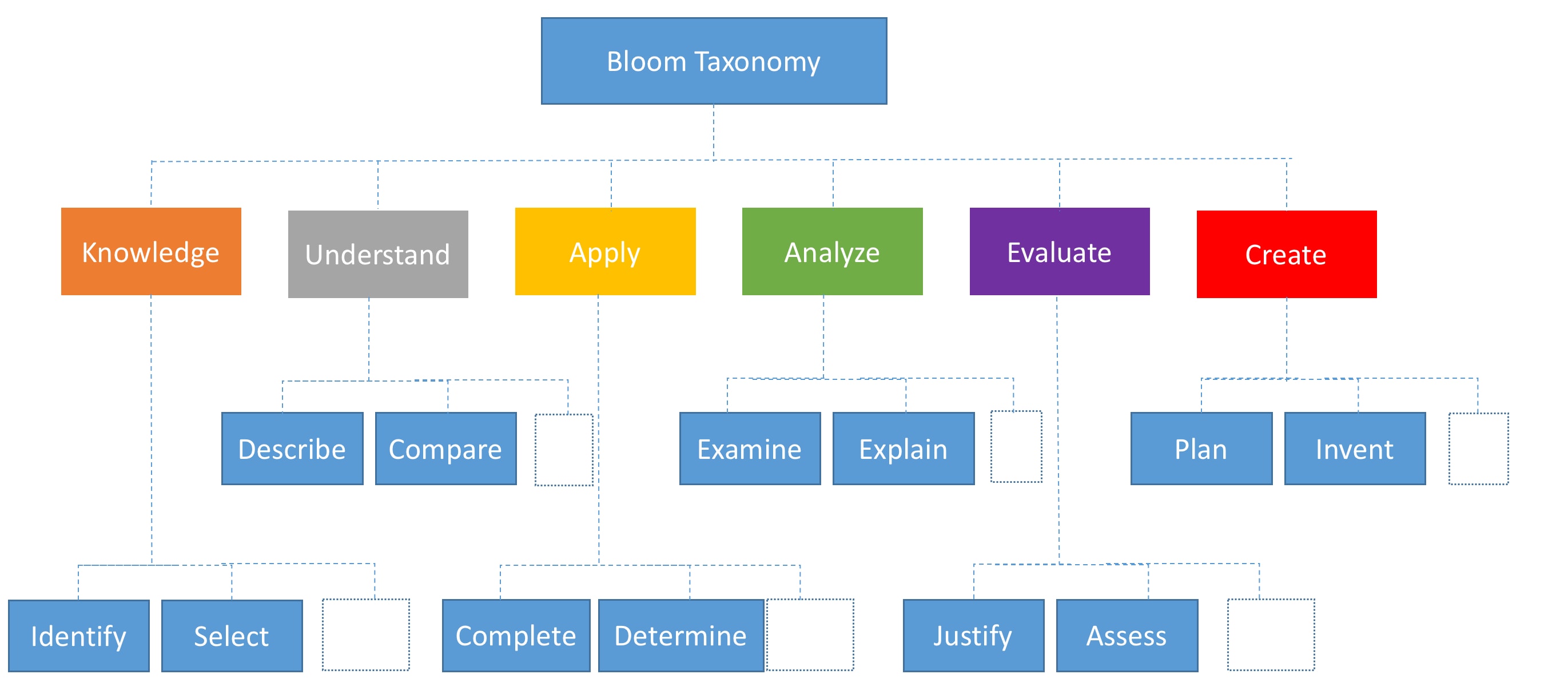}
\caption{A representative taxonomy of Bloom's verbs}
\label{fig:BloomTaxonomy}
\end{figure}

\figref{BloomTaxonomy} shows a representative view of Bloom's taxonomy. For example, the verb \textit{knowledge} has a list of child verbs such as \textit{identify} and \textit{select}. Similarly, other top-level verbs have their own set of verbs. We experiment with a subset of $10$ verbs, as recommended by SMEs. We also explore another more condensed list as suggested by the same SMEs to investigate the potential of hierarchical options. We collapse the 10 verbs belonging to the same parent, resulting in 4 higher-level verb classes in Bloom's taxonomy. The verb classes used in our experiments are given in \tabref{bloom-verb-list}.

\begin{table}[!t]
\footnotesize
\begin{center}
\begin{tabular}{cccc}
\toprule
& & \multicolumn{2}{c}{\textbf{Distribution}} \\
\textbf{Original List} & \textbf{Collapsed List} & \bank & \pharma \\
\midrule
identify & \multirow{6}{*}{knowledge} & 542 & 323 \\
define &  & 85 & 12 \\
recall &  & 36 & 11 \\
recognize &  & 35 & 31 \\
select &  & 6 & 1 \\
list &  & 1 & 8 \\
\hdashline
describe & \multirow{2}{*}{understand} & 144 & 166 \\
explain &  & 127 & 65 \\
\hdashline
outline & analyze & 11 & 40 \\
\hdashline
determine & apply & 5 & 5 \\
\bottomrule
\end{tabular}
\end{center}
\caption{Bloom's verbs used for generating Learning Objectives and their distribution from a random sample of 100 chunks. Each chunk often has more than one keyphrase describing it, requiring the SMEs to suggest a matching Bloom's verb.}
\label{tab:bloom-verb-list}
\end{table}

To associate a verb from Bloom's taxonomy with a keyphrase learning objective, we train a multilayer perceptron (MLP) to predict a verb given a document (or chunk) and a candidate keyphrase. Thus, the MLP consists of two fully connected (dense) layers with ReLU activation functions\cite{ReLU} in each node. The input of the network is the mean bag-of-words embedding of the document text and the keyphrase. 


Word embeddings are pre-trained \wordtovec embeddings \cite{Mikolov+:2013a,Mikolov+:2013b} trained on the English Wikipedia. Word embeddings are kept static and not updated during back-propagation.\footnote{We also experimented with updating the embeddings (Facebook's \texttt{fastText}), but found little improvement and thus chose the simpler static model with fewer parameters.} 
This approach of predicting bloom verbs was found to be very effective as shown in Section~\ref{sec:exp:lo}. 


Two examples of generating learning objectives are shown in \tabref{bloomKPExamples}. They show the pairing of a Bloom's verb with various keyphrases. These pairings are presented to SMEs to evaluate, where their ratings allow us to determine the final rankings to select the most appropriate candidates as learning objectives for a piece of text. Note that the text in the examples (from a document chunk) has been truncated for presentation.

\begin{table}[!t]
\centering
\footnotesize
\begin{tabular}{|l|l|l|}
\hline
\textbf{Bloom's} & \textbf{Keyphrase} & \textbf{Avg.} \\
 Verb &  & \textbf{Rating} \\
\hline
describe & ach payments & 3 \\
explain & ach transaction flow & 2.5 \\
describe & ACH transactions & 2.5 \\
identify & ACH network & 2 \\
identify & ACH networks & 2 \\
identify & ACH payment request & 2 \\
describe & ACH payments industry & 2 \\
explain & internal ach transaction & 2 \\
identify & traditional ACH payments & 2 \\
identify & ACH & 1 \\
\hline
\multicolumn{3}{|c|}{\textbf{Text}}\\
\hline
\multicolumn{3}{|p{\linewidth}|}{ACH Payments In this section we are going to take a look at a payment type generically known as small value  electronic credit transfers, although they are referred to with a number of different names,  including automated clearing house or ACH transactions, automatic clearing payments,  electronic clearing payments and giro payments. \dots } \\
\hline
\hline
\textbf{Bloom's} & \textbf{Keyphrase} & \textbf{Avg.} \\
 Verb &  & \textbf{Rating} \\
\hline
explain & consumer payments & 3 \\
define & Large value payments & 2 \\
describe & payments industry & 2 \\
define & Small value payments & 2 \\
identify & consumer bill payments & 1.5 \\
recall & consumer payments operations & 1.5 \\
identify & corporate-to-corporate payments & 1.5 \\
identify & interbank payments & 1.5 \\
explain & payments & 1.5 \\
identify & banks & 1 \\
\hline
\multicolumn{3}{|c|}{\textbf{Text}}\\
\hline
\multicolumn{3}{|p{\linewidth}|}{Business Overview Why focus on consumer payments? There are two sides to this question.  First, why do banks focus on consumer payments? There are several reasons: Banks cannot accept consumer deposits without providing payment services linked to  those accounts. While consumer deposits have always been important, they have never  been as important as they are today. \dots }\\
\hline
\end{tabular}
\caption{Examples of generating learning objectives and their average ratings from SMEs.}
\label{tab:bloomKPExamples}
\end{table}




\section{Experiments}




\subsection{Data sets}

We evaluate our chunking and learning objective systems on real-world documents from two industries: banking and finance (henceforth \bank) and pharmaceuticals (henceforth \pharma).
\tabref{data:stats} summarizes the word statistics of the two document collections used in our experiments.

The \bank data set serves as our initial dataset for tuning and testing our methodology, which has a mix of 15 ``formal'' (e.g.\ Microsoft Word style) documents and 15 ``informal'' (e.g.\  HTML, MediaWiki style, Microsoft PowerPoint slides) documents.


The \pharma data is a set of client-provided documents with a similar distinction of formal and informal documents. The \pharma data set consists of 382 courses containing 408 documents, where most courses only have one document. We develop our methodology on the \bank data set and pursue a deployment on the \pharma data set (detailed in \secref{deployment}). The remainder of this section describes our experimental results on the \bank data set.


\subsection{Evaluation: Document Chunking}
\label{sec:exp:chunking}

For tuning and evaluation, we require gold standard chunks for the \bank documents. To this end, we ask SMEs to chunk\footnote{Chunks are contiguous breaks in the document, so chunk boundaries can be succinctly described and compared using the starting line/page number for each chunk.} these documents manually, resulting in $243$ chunks in total for the $30$ documents.
The documents were chunked by SMEs (with inter-annotator disagreements of the chunk boundaries resolved) based on their understanding of the subject from an instructional design perspective. The SMEs opted for page level chunks and thus we build our measure of quality at the page level.





To measure the quality of our system against SMEs, we compute the average F1 score on their list of chunk boundaries. We omit the first chunk boundary as it always starts at page 1, and penalise duplicate page numbers (i.e. multiple sections on the same page). To illustrate the evaluation method, we give an example:
\begin{align*}
\text{system chunks} &= [ 1, 4, 4] \\
\text{human chunks}  &= [ 1, 3, 4 ]
\end{align*}
where each number in the list denotes the starting page number of a chunk. We omit the first chunk, yielding:
\begin{align*}
\text{system chunks} &= [4, 4]\\
\text{human chunks} &= [3, 4]
\end{align*}

Precision of the system is therefore $1/2 = 0.5$ (the second starting page number ``4'' is penalised), the recall is $1/2 = 0.5$, and thus F1 $=0.5$.




There are a number of hyper-parameters for our chunking methods, which are available in Tables~\ref{tab:hyperparams:syntactic} and~\ref{tab:hyperparams:semantic}. 
We tune them manually based on the F1 score using a small labeled development set. 
Given the tuned models, we apply them to the \bank documents.


\begin{table}[!t]
\footnotesize{
\begin{center}
\begin{tabular}{rcc}
\toprule
& \textbf{\bank} & \textbf{\pharma} \\
\midrule
No. Documents & 30 & 408 \\
No. Word Tokens & 376,570 & 1,251,712 \\
Vocabulary Size & 32,598 & 92,890 \\
\bottomrule
\end{tabular}
\end{center}}
\caption{Data set statistics.}
\label{tab:data:stats}
\end{table}

From the chunking performance in \tabref{results:chunking}, we found that for formal documents, the \syntactic (relying on the font size to detect natural chunk boundaries) has the highest accuracy for formal content. In contrast, for the informal content, where structural information may not be very indicative of natural chunk boundaries, we find that the \semantic gives better results as expected.  

In order to qualitatively assess the results of our systems, we also evaluate them with a blind user study. Two expert instructional designers were presented the output of chunks by different chunking algorithms in random order and without information on the underlying algorithm. Each designer was asked to rate a chunk output with 1 (poor), 2 (acceptable), and 3 (good) based on their quality and usefulness from an Instructional Design point of view. Due to complexity and unsupervised nature of the task, ratings above 1 are strongly encouraging.

As seen in \tabref{results:chunking}, the average ratings for all our best systems is greater than $1.5$ indicating our system generated chunks could be acceptable and useful for instructional designers. Furthermore, we find that the scores from the user study reinforce the assessment that formal content (with well structured natural chunk boundaries) are reliably chunked using the \syntactic algorithm while informal content is better chunked using the \semantic algorithm.

Surprisingly, we find that the \hybrid chunking algorithm performs poorly on informal content compared to the \semantic. However, the average user evaluation rating shows that the resulting chunks are highly acceptable, as expected from initial trials in designing this algorithm. Our inspection shows that increasingly the granularity from lines to font groups simply means the desired chunk boundaries are often missed (and they are near misses), and that fewer chunks are created. We reason that fewer chunks are favorable to users when the document does not have clear chunking boundaries because of simplicity. Furthermore, our F1-score measure is strict, meaning near misses for chunk boundaries are also heavily penalized, but the chunk boundaries of the \hybrid algorithm may be acceptable to the user. We also experimented with alternative methods such as repositioning the chunk start indices from the \semantic to match boundaries given by the \syntactic, but the resulting chunks were not favored by the SMEs in initial trials.

Overall, the \syntactic performs well on both formal and informal documents for the \bank data set. On inspection of the informal documents, some contain sufficient structure for the \syntactic to infer the desired chunking boundaries, whereas documents with non-usable structures, the \semantic provides more favorable chunking boundaries. We also reason that the higher ratings for the \syntactic is due to the \syntactic finding section headings for chunking boundaries, which seems to be preferred by users, whereas another grouping of pages for the chunk may be more appropriate. These chunking systems provide variety, ensuring that we have a suitable set of chunks for any document.


\begin{table}[!t]
\footnotesize{
\begin{center}
\begin{tabular}{cccc}
\toprule

\textbf{System} & \textbf{Doc Type} & \textbf{F1} & \textbf{Avg. Rating}\\
\midrule
\multirow{3}{*}{\parbox{1.85cm}{\centering\method{syntactic-}\method{chunker}}}
& Formal & {\bf 0.62} & {\bf 2.17} \\
& Informal & 0.31 & 2.00 \\
& Combined & 0.47 & 2.08 \\
\hdashline
\multirow{3}{*}{\parbox{1.85cm}{\centering\method{semantic-}\method{chunker}}}
& Formal & 0.08 & 1.36 \\
& Informal & {\bf 0.20} & {\bf 1.67} \\
& Combined & 0.14 & 1.51 \\
\hdashline
\multirow{3}{*}{\parbox{1.85cm}{\centering\method{hybrid-}\method{chunker}}}
& Formal & {\bf 0.21} & 1.49 \\
& Informal & 0.05 & {\bf 1.77} \\
& Combined & 0.13 & 1.63 \\
\bottomrule
\end{tabular}
\end{center}}
\caption{Results for Document Chunking on the \bank data set. Bold values indicate the best performance for that system.}
\label{tab:results:chunking}
\end{table}








\subsection{Evaluation: Learning Objective Generation} \label{sec:exp:lo}

To collect annotation for evaluation and for training the Bloom's verb MLP and for keyphrase selection, we present to SMEs: a document chunk (manually chunked by different SMEs in \secref{exp:chunking}) and the top-10 NLU generated keyphrases and ask them to (1) rate the keyphrase in terms of usefulness as a learning objective suffix on an ordinal scale from 1--3 (same as chunking evaluation) and (2) select an appropriate Bloom's verb (out of 10 verbs) for the particular keyphrase.

We randomly sample from the full $243$ document chunks and collect annotations for 100 chunks, where each chunk is annotated by 2 SMEs. We aggregate these keyphrase ratings by taking the mean rating. For Bloom's verb selection, we ask the judges to agree on a particular verb if there is discrepancy. To generate gold standard for the condensed verbs ($4$ classes), we map the original 10 classes to the $4$ classes, as given in \tabref{bloom-verb-list}. 



\begin{table}[!t]
\footnotesize{
\begin{center}
\begin{tabular}{rccc}
\toprule
 & \textbf{P@1} & \textbf{P@3} &\textbf{P@5} \\
\midrule
\footnotesize
{\bf Avg. Rating} & 1.97  & {\bf 2.23} & {2.20} \\
{\bf Precision} & {0.5} & {\bf 0.5}& {0.45} \\
\bottomrule
\end{tabular}
\end{center}}
\caption{{\textbf{\bank}}: Candidate Keyphrase Selection for Learning Objective Generation}
\label{tab:KP-rank-bank}
\end{table}

\subsubsection{Candidate Keyphrase Selection}

We use 10-fold cross-validation at the chunk level for our experiments. We select the top-$k$ keyphrases for each chunk as candidates for the learning objectives of that chunk. From Equation~\ref{eq:maxKPAverage}, the tuning of factor weights is based on the average user rating of these top-$k$ keyphrases. 

We evaluate the quality of candidate keyphrase selection using the average user rating of the selected keyphrases, and Precision@N defined as
\begin{equation}
P@N = \frac{k_g \cap k_s}{|k_s|}
\end{equation}
where $k_g$ is the set of gold standard keyphrases that have an average user rating of at least $1.5$\footnote{We want our system to select only good quality keyphrases.}, and $k_s$ is the set of top-$k$ keyphrases selected by the system. This measure shows whether our selection methods are returning the keyphrases that are relevant for each chunk as determined by the SMEs.

From \tabref{KP-rank-bank} our keyphrase selection method has a P@5 of $0.45$ with a high average user rating. This means that 45\% of the top 5 keyphrases selected contain the gold standard keyphrases. 

\subsubsection{Selecting Bloom's Verbs}

Given a document and its verbs from the Bloom taxonomy, we train an MLP and optimise its hyperparameters 
based on 10-fold cross-validation at the chunk level. We use the evaluation metric of mean F1 score over the 10-folds.\footnote{For a particular fold, we compute weighted F1, where it is weighted by the number of true instances for each class.}  We use 2 test sets: (1) all keyphrases and (2) top-5 keyphrases predicted by our system. Note that in each fold, the training data remains the same, but test set (2) is a subset of (1).

We present the classification performance of Bloom's verbs in \tabref{bloom-bank}. As expected, the performance in the 4-class prediction task is better than the 10-class prediction due to less confusion amongst classes. Baseline experiments where we assign the majority class for all predictions show a consistent 0.10 drop in F1-score for both the 4-class and 10-class prediction scores. 





\begin{table}[!t]
\footnotesize{
\begin{center}
\begin{tabular}{ccc}
\toprule
\multirow{2}{*}{\textbf{Test Set}} & \multicolumn{2}{c}{\textbf{F1}} \\
&\textbf{4-Class} & \textbf{10-Class}\\
\midrule

All KP & 0.69 & 0.51 \\
System Top-5 KP & 0.70 & 0.53 \\

\bottomrule
\end{tabular}
\end{center} }
\caption{{\textbf{\bank}}: Bloom's verb (BV) prediction performance. ``KP'' denotes keyphrase.}
\label{tab:bloom-bank}
\end{table}





\vfill

\section{Deployment} \label{sec:deployment}

Making content discoverable is a key challenge faced by talent development teams in organizations worldwide. Our system addresses this challenge and is currently being piloted at one of the world's largest pharmaceutical companies to help organize their learning content.

\noindent {\bf Experiments and Tuning: } Using the \pharma data shared by the pharmaceutical company (statistics in \tabref{data:stats}), we repeated the \bank data set experiments on this data. The pharmaceutical SMEs only wanted generation of document level learning objectives, and not document chunking. Thus, we describe only the experiments for this task. As with the \bank experiments, we ask SMEs to rate predicted keyphrases and select the appropriate verb (from the Bloom's taxonomy) given a document\footnote{These were the original documents and were not chunked.} and keyphrase.\footnote{We collect annotations for a random 25\% subset of the $408$ (original) documents, as SMEs simply did not have the time to evaluate all documents due to their length.} 
For learning objective keyphrase selection and Bloom's verb prediction, we train and tune the systems with 10-fold cross-validation as before.

\begin{table}[!t]
\footnotesize{
\begin{center}
\begin{tabular}{rccc}
\toprule
 & \textbf{P@1} & \textbf{P@3} &\textbf{P@5} \\
\midrule
\footnotesize
{\bf Avg. Rating} & 1.24 & 1.35 & {\bf 1.38}  \\
{\bf Precision}  & {0.1} & 0.3& {\bf 0.32}  \\

\bottomrule
\end{tabular}
\end{center}}
\caption{\textbf{\pharma}: Candidate Keyphrase Selection for Learning Objective Generation}
\label{tab:KP-rank-pharma}
\end{table}

Keyphrase selection and Bloom's verb prediction performance for \pharma are presented in \tabref{KP-rank-pharma} and \tabref{bloom-pharma}. We find that our candidate keyphrase average rating and precision is lower than what was seen for banking data. We hypothesize a reason for this is due to the extremely dense and domain specific content as well as the requirement of complete documents without chunking when generating learning objectives.

Furthermore, many documents from the pharmaceutical company refer to chemical compounds and chemical formulae, which resulted in skewed TF-IDF weights while selecting candidate keyphrases. Our hypothesis is also backed by the score weights for TF-IDF become less important for \pharma data as compared to \bank data. We note that the Google N-grams scores were useful for re-ranking keyphrases in both domains. The results also suggest that domain-specific adaption of keyphrase extraction methods (eg. supervised methods) may be required for learning objective generation in content that is very technical.

For Bloom's verb prediction (\tabref{bloom-pharma}), we see a marginally lower performance, but the trend largely remains the same.

\begin{table}[!t]
\footnotesize{
\begin{center}
\begin{tabular}{ccc}
\toprule
\multirow{2}{*}{\textbf{Test Set}} & \multicolumn{2}{c}{\textbf{F1}} \\
&\textbf{4-Class} & \textbf{10-Class}\\
\midrule
All KP & 0.66 & 0.50 \\
Top-5 System KP & 0.71 & 0.48 \\
\bottomrule
\end{tabular}
\end{center}}
\caption{{\textbf{\pharma}}: Bloom verb prediction performance. ``KP'' denotes keyphrase.}
\label{tab:bloom-pharma}
\end{table}

\begin{table}[!t]
\footnotesize{
\begin{center}
\begin{tabular}{ccc}
\toprule
\multirow{3}{*}{\textbf{System}} & \multicolumn{2}{c}{\textbf{Avg. Time Per}}\\
 & \multicolumn{2}{c}{\textbf{Document (seconds)}}\\
 & \textbf{\bank} & \textbf{\pharma} \\
\midrule
\syntactic & 0.41 & 0.20 \\
\semantic & 0.40 & 0.20 \\
\hybrid & 0.49 & 0.27 \\
\keyphrase & 0.02 & 0.02 \\
\learningobjective & 0.03 & 0.02 \\
\bloom & 0.05 & 0.04 \\
\bottomrule
\end{tabular}
\end{center}}
\caption{Throughput: Document Chunking, Keyphrase generation,  candidate keyphrase selection, and bloom verb prediction (in seconds)}
\label{tab:throughput:chunking}
\end{table}


\subsection{Commercial Deployment}

\begin{figure*}[!t]
\centering
\includegraphics[width=0.8\linewidth]{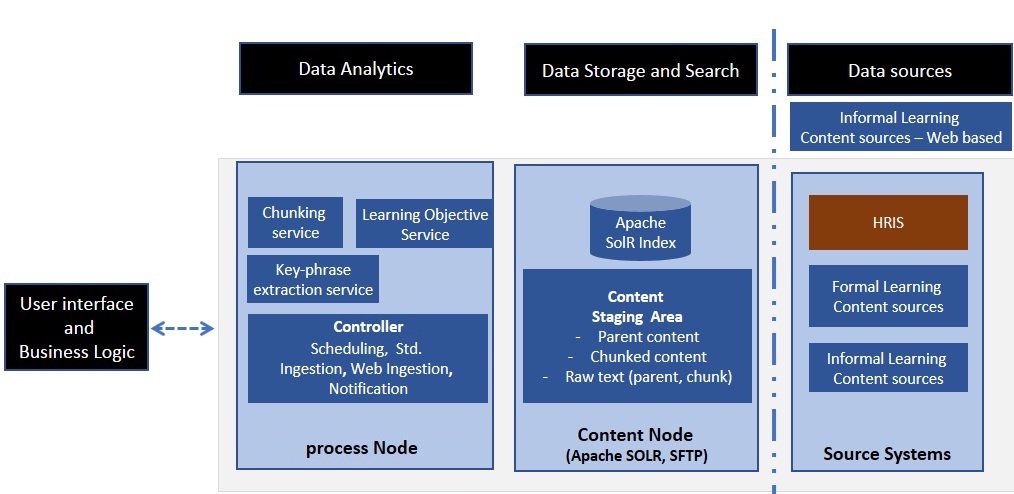}
\caption{High-level architecture diagram}
\label{fig:architecture}
\end{figure*}


A collection of over $20,000$ learning courses have been labeled with learning objectives generated by our system and are being imported into existing learning management systems used by the organization. This is to help the organization retrieve courses efficiently, identify similar course material and prioritize new course development as it allows them identify gaps in their course material by checking course objectives not covered existing in course material. We briefly describe the architecture of our full system as this is the eventual deployment goal.


\subsection{System Architecture} 

Broadly, the system consists of three subsystems (see Figure \ref{fig:architecture}): (1) {\bf UI and Business logic layer}, which exposes interfaces for search and enforces business logic for user access; (2) {\bf Data Analytics layer}, which are Web services for document chunking, keyphrase extraction, learning objective generation. Additional web services that generate different metadata can be easily plugged in and integrated into our system; and (3) {\bf Data Storage and Search}, where we use Apache Solr to store all generated metadata and document text and to enable search. An illustration of the architecture is presented in \figref{architecture}. Physical documents can either be stored locally or can be accessed via remote requests to learning management systems. Data ingestion from formal course repositories as well as informal sources (web based or Intranet) are supported. 


We use document format specific APIs to physically persist document chunks in their original file formats. Our system exposes a simple search interface by which users can query the system using learning objectives. The system allows refinement of search results and also defines user workspaces where course packages can be created and shared. 



Table~\ref{tab:throughput:chunking} summarizes the average throughput for each of our components (computed on an Intel i5 6300 2.4 Ghz CPU with 8 GB RAM), demonstrating its speed and ease of scalability for large scale processing.


\section{Discussion and Conclusion} \label{sec:discussion}

In this paper, we presented the first system that automatically chunks learning material and generates learning objectives derived from content. It consists of modular sub-components that require little training data for adaptation. The cloud based web service architecture enables effective use of each of its capabilities. 

Our system uses a state-of-the-art embedding-based approach to chunk learning material into meaningful chunks. It also uses generic structural features from the document to guide chunking. It employs a novel methodology for generating learning objectives, which combines automatically generated verbs from Bloom's taxonomy and extracted keyphrases.

Our system's capabilities are being used by a large pharmaceutical company to organize learning material. We present detailed experiments on two different domains that demonstrate the applicability of our work. 

In future work, we look to extend the work with improvements to our document ingestion capabilities, such as supporting images and videos using OCR and extracting headers and footers, and tabulated data. We would also like to add capabilities that aid instructional designers with other aspects of course design, such as discovering similar courses, summarizing documents, and improving learning objective generation to support a wider set of verbs from Bloom's taxonomy as well as supervised approaches for keyphrase generation in highly technical domains.



\section*{Acknowledgements}

We thank our colleagues from IBM Global Business Services for leading this project on the business side. In particular, many thanks to Prasanna C Nair, Sandra Misiaszek, \mbox{Madhusmita} P Patil, Narasimhan K Iyengar, Renjith K Mathew, Partha S Guha, Pinaki Chakladar, Richa Sethi, Tulasi S Manepalli, Anindita Gupta, and Vinod Uniyal. Our research would not have been possible without their vision, support, data, expertise, and client engagements.


\bibliographystyle{siam}
\bibliography{edm2018}

\end{document}